\documentclass[11pt]{article}
\usepackage{acl2012}
\usepackage{times,latexsym,multirow,url,booktabs,fixltx2e,amssymb,amsmath,subfig,tabularx,colortbl,color,graphicx,setspace}
\DeclareMathOperator*{\argmax}{arg\,max}
\definecolor{Gray}{rgb}{0.68,0.68,0.68}
\newcommand{\specialcell}[2][c]{\begin{tabular}[#1]{@{}l@{}}#2\end{tabular}}

\newcommand{\eat}[1]{}

\def\papernumber #1 raised #2 {
  \vspace{-#2}
  \vbox to 0pt{\framebox{\bf Paper Number: #1}}
}

\newcommand{\smgap}{-5pt}

\newcommand{\secref}[1]{Section~\ref{#1}} 

\newcommand{\hide}[1]{} 

\title{KERT: Automatic Extraction and Ranking of Topical Keyphrases from Content-Representative Document Titles}

\author{Marina Danilevsky\textsuperscript{1}, Chi Wang\textsuperscript{1}, Nihit Desai\textsuperscript{1}, Jingyi Guo\textsuperscript{2}, and Jiawei Han\textsuperscript{1} \\
\textsuperscript{1}Dept of Computer Science, University of Illinois at Urbana-Champaign\\
\textsuperscript{2}Dept of Computer Science, University of Massachussets Amherst\\
 {\tt \footnotesize{\{danilev1,chiwang1,nhdesai2\}@illinois.edu, jingyi@cs.umass.edu, hanj@cs.uiuc.edu}} }

\date{}

\begin{document}
\maketitle
\begin{abstract}

We introduce KERT (Keyphrase Extraction and Ranking by Topic), a framework for topical keyphrase generation and ranking. By shifting from the unigram-centric traditional methods of unsupervised keyphrase extraction to a phrase-centric approach, we are able to directly compare and rank phrases of different lengths. We construct a topical keyphrase ranking function which implements the four criteria that represent high quality topical keyphrases (coverage, purity, phraseness, and completeness). The effectiveness of our approach is demonstrated on two collections of content- representative titles in the domains of Computer Science and Physics.

\end{abstract}

\setstretch{0.97}


\vspace{-0.1cm}
\section{Introduction}\label{sec:intro}

Keyphrases have traditionally been defined as terms or phrases which summarize the topics in a document \cite{Turney00}. Keyphrase extraction is an important step in many tasks, such as document summarization, clustering, and categorization \cite{Manning99}. More recently, the definition has been expanded to include the notion of topical keyphrases - groups of keyphrases which summarize the topics in a given document, or document collection \cite{Liu10}.
Most existing work on keyphrase extraction identifies keyphrases from either individual documents or an entire text collection \cite{Tomokiyo03,Liu10}. However, recently there has been some interest in working with documents consisting of very short text, such as a collection of tweets \cite{Zhao11}, in order to summarize the document collection.

Our framework also targets short texts - in particular, collections of \emph{content-representative titles}. Document titles are content-representative if they may serve as concise descriptions of the content of the document. The words in a content-representative title can therefore be thought of as probabilistic priors for which words are the most likely to generate keyphrases describing the document. Scientific publications and newspaper articles generally have content-representative titles, whereas fiction books generally do not. As we address the task of representing the topics present in a document collection, content-representative titles are cleaner and more efficient to deal with than entire documents\hide{, and we successfully address this task}.




Most current approaches to topic construction yield lists of unigrams to represent topics. However, it has long been known that unigrams account for only a small fraction of human-assigned index terms \cite{Turney00}. Therefore, in order to construct high quality keyphrases for a given topic, it is important to provide n-gram keyphrases rather than unigram keywords. However, it is inappropriate to discard all unigrams when approaching this task. For instance, consider that the unigram `classification' and the trigram `support vector machines' are both high quality topical keyphrases for the Machine Learning topic in the domain of Computer Science. We should therefore be able to perform integrated ranking of mixed-length phrases in a natural way.


Such a ranking function must successfully represent human intuition for judging what constitutes a high quality topical keyphrase. We propose that this is best represented by four criteria: coverage, purity, phraseness, and completeness. As an example, consider the task of constructing and ranking keyphrases for various topics in Computer Science:\\
$\bullet$ \textbf{Coverage:} A representative keyphrase for a topic should cover many documents within that topic. \emph{Example: `information retrieval' has better coverage than `cross-language information retrieval' in the Information Retrieval topic.}\\
$\bullet$ \textbf{Purity:} A phrase is pure in a topic if it is only frequent in documents belonging to that topic and not frequent in documents within other topics. \emph{Example: `query processing' is more pure than `query' in the Database topic.}\\
$\bullet$ \textbf{Phraseness:} A group of words should be combined together as a phrase if they co-occur significantly more often than the expected chance co-occurrence frequency, given that each term in the phrase occurs independently. \emph{Example: `active learning' is a better phrase than `learning classification' in the Machine Learning topic.}\\
$\bullet$ \textbf{Completeness:} A phrase is not complete if it is a subset of a longer phrase, in the sense that it rarely occurs in a title without the presence of the longer phrase. \emph{Example: `support vector machines' is a complete phrase, whereas `vector machines' is not because `vector machines' is almost always accompanied by `support'.}

Our aim is to construct topics represented by a ranked list of keyphrases of various lengths where more highly ranked keyphrases are considered to be better index phrases for that topic. The main contributions of this work are:
\\
$\bullet$ We introduce KERT (Keyphrase Extraction and Ranking by Topic), a framework for topical keyphrase generation and ranking. By altering the steps in the traditional methods of unsupervised keyphrase extraction,  we can directly compare of phrases of different lengths, resulting in a natural integrated ranking of mixed-length keyphrases.\\
$\bullet$ We construct a topical keyphrase ranking function which implements the four criteria that intuitively represent high quality topical keyphrases (coverage, purity, phraseness, and completeness).\\
$\bullet$ We demonstrate the effectiveness of our approach on two collections of content-representative titles in the domains of Computer Science and Physics.

\hide{The rest of this paper is organized as follows: We first review the related literature in Section~\ref{sec:relwork}. Section~\ref{sec:method} presents our methodology and framework, and is followed by Section ~\ref{sec:exp}, which presents experimental results on two real world datasets. Section~\ref{sec:conc} concludes and discusses directions for future work.}


\vspace{-0.1cm}
\section{Related Work}\label{sec:relwork}
\vspace{-0.1cm}


The state-of-the-art approaches to unsupervised keyphrase extraction have generally been graph-based, unigram-centric ranking methods, which first extract unigrams from text, rank them, and finally combine them into keyphrases. TextRank~\cite{Mihalcea04} constructs keyphrases from the top ranked unigrams in a document collection. Topical PageRank \cite{Liu10} splits the documents into topics and creates keyphrases from top ranked topical unigrams.
Some previous methods have used clustering techniques on word graphs for keyphrase extraction \cite{Liu09,Grineva09}, relying on external knowledge bases such as Wikipedia to calculate term importance and relatedness. Barker and Cornacchia \shortcite{Barker00} use natural language processing techniques to select noun phrases as keyphrases.
Tomokiyo and Hurst \shortcite{Tomokiyo03} take a language modeling approach, requiring a document collection with known topics as input and training a language model to define their ranking criteria.

Unlike most of these methods which extract keyphrases from documents, we aim to extract keyphrases from a corpus of short texts. Zhao et al \shortcite{Zhao11} also uses short text - microblogging data from Twitter - but we work with content-representative document titles. 

Topic modeling techniques such as PLSA (probabilistic latent semantic analysis) \cite{Hofmann01} and LDA (latent Dirichlet allocation) \cite{Blei03} take documents as input, model them as mixtures of different topics, and discover word distributions for each topic. 
Some previous work has developed topic modeling to discover topical phrases comprised of consecutive words \cite{Wang07}. Our framework also uses topic modeling - an extension of LDA - to perform the initial word clustering into topics, but does not restrict phrases to only those word sequences explicitly found in the text.

Since we aim to transform a collection of short texts into sets of ranked keyphrases, the top-K keyphrases for each topic may also serve as that topic's labels. Therefore our work is tangentially related to automatic topic labeling \cite{Mei07}.


\section{KERT Framework}\label{sec:method}
\vspace{-0.1cm}

A key aspect of our framework is that we do not follow the traditional unigram-centric approach of keyphrase extraction, where words are first extracted and ranked independently, and then combined to create phrases. Instead, we construct topical phrases immediately after clustering the unigrams. By shifting from a unigram-centric to a phrase-centric approach, we are able to extract topical keyphrases and implement a ranking function that can directly compare keyphrases of different lengths, as we explain in Section~\ref{sec:rank}. \hide{We thus avoid the pitfall of combining two high-ranked unigrams into a bad keyphrase - e.g., `classification' and `learning' would rank highly in the Machine Learning topic, but the phrase `learning classification' is not meaningful. Note that we do not only consider consecutively appeared words to be phrases because people may insert words between meaningful phrases when making titles. And we do not use syntactic approaches to parse the text. }

Our steps for topical keyphrase generation and ranking are as follows:

\textbf{Step 1.} Cluster words in the document dataset into T foreground topics and one background topic, using background LDA. 

\textbf{Step 2.} Extract candidate keyphrases for each topic according to the word topic assignment. 

\textbf{Step 3.} Rank the keyphrases in each foreground topic $z\in1,\ldots,k$ by integrating the criteria of Coverage, Purity, Phraseness, and Completeness.

In the subsequent sections we give detailed explanations of \hide{the implementation of }each step\hide{ and the measures that were constructed}.

\subsection{Clustering Words using Topic Modeling}
\label{sec:clustering}
Content-representative titles exhibit several characteristics that  guide our clustering approach:
\\
$\bullet$ \textbf{Short and well-formed.} Most titles are composed of a few key technical words and background words. Background words are found across titles belonging to different topics.\\
$\bullet$ \textbf{Ambiguous Unigrams\hide{Unambiguous Phrases}.} A single word (e.g. `vector') may appear in different topics, while phrases (e.g. `support vector machines') are overall less topic-ambiguous.\\
$\bullet$ \textbf{Mix of Topics.} A title may comprise a sparse mixture of different topics, in spite of its short length.\hide{\footnote{We thus relax the constraint seen in \cite{Zhao11} that each short text may only belong to one topic, since scientific papers or newspaper articles often span multiple topics.}}

LDA \cite{Blei03} and its extensions have been shown to be effective for modeling textual documents. We therefore use a modified LDA model which includes an additional background topic $z=0$. Each title is modeled by a distribution over the foreground topics $z=1,\dots,k$ and the background topic. For each word in a title, we decide if it belongs to the background topic or one of the foreground topics, and then choose the word from the appropriate distribution.

Formally, let $\phi^t$ denote the word distribution for topic $t=0,\dots,k$. Let $\theta^d$ be the topic distribution of document $d$. Let $\lambda$ denote a Bernoulli distribution that chooses between the background topic and foreground topics. The generative process is as follows:

\begin{enumerate}
\item Draw $\phi^t\sim Dir(\beta)$, for $t=0,\ldots,k$.
\item For each title $d\in D$,

\begin{enumerate}
 \item draw $\theta^d\sim Dir(\alpha)$.
\item for each word position $i$ in $d$
\begin{enumerate}
\item draw $y_{d,i}\sim Bernoulli(\lambda)$
\item if $y_{d,i}=0$, draw $w_{d,i}\sim Multi(\phi^0)$, otherwise
\begin{enumerate}
\item draw topic $z\sim \theta^d$
\item draw $w_{d,i} \sim Multi(\phi^z)$
\end{enumerate}
\end{enumerate}
\end{enumerate}
\end{enumerate}
where $\alpha$ and $\beta$ are Dirichlet prior parameters for $\theta$ and $\phi$ respectively.


We use a collapsed Gibbs sampling method for model inference. We iteratively sample the topic assignment $z_{d,i}$ for every word $w_{d,i}$ in each document until convergence. In traditional topic modeling tasks, the sampled topic assignments are mainly used to estimate the topic distribution $\phi^z$. In our case, we are more interested in the topic assignments for the words in each title, because these values are the foundation of our topical keyphrase generation step, as described in the next section. We use the \emph{maximum a posteriori} (MAP) principle to label each word: $z_{d,i}=\argmax_{z_{d,i}=0,\ldots,k} P(z_{d,i}|W)$.

\subsection{Candidate Keyphrase Generation}
\label{sec:mining}

There are two ways to discover keyphrases: either extract them from the text (sequences of words) which actually occur in the text, or to automatically construct them \cite{Frank99}, an approach which is regarded as both more intelligent and more difficult \cite{Hammouda05,Manning99}.

In a dataset of content-representative titles, extracting phrases directly from the text is quite limiting as this approach is too sensitive to the caprices of various writing styles. For instance, consider that two computer science papers titles, one containing `mining frequent patterns' and the other containing `frequent pattern mining,' are clearly discussing the same topic, and should be treated as such. A keyphrase may also be separated by other words: `\textbf{mining} top-k \textbf{frequent} closed \textbf{patterns}' also belongs to the topic of frequent pattern mining, in addition to incorporating secondary topics of top-k frequent patterns, and closed patterns. Therefore, we define a phrase to be an order-free set of words appearing in the same title, and must therefore \emph{construct} our phrases.

After the clustering step described in Section \ref{sec:clustering} is completed, each word $w_{d,i}$ in each title $d$ has a topic label $z_{d,i}\in \{0,\dots,k\}$. If a set of words in a title $d$ are assigned a common foreground topic $t>0$, these words may comprise a topical phrase in $t$ (e.g. \{`frequent',`pattern',`mining'\} in the topic of `Data Mining'). If this set occurs in many titles with the topic label $t$, it is likely a good candidate keyphrase for that topic.  

We use frequent pattern mining approaches to mine the candidate topical phrases. For each topic $t$, we construct a topic-$t$ word set $p^t_{d}=\{w_{d,i}|z_{d,i}=t\}$ consisting of the words with the topic label $t$ for each document $d$. We unite all the topic-$t$ word sets into the topic-$t$ set $D_t=\{d|p^t_{d}\neq \emptyset\}$. We may then mine frequent word sets from $p^t_{D_t}=\{p^t_d|d\in D_t\}$ using any efficient pattern mining algorithm, such as FP-growth \cite{Han04}. We require a good topical keyphrase to have enough topical support $f_t(p)>\mu$ in order to filter out some coincidental co-occurrences (where $f_t(p)$ denotes the frequency of the word set $p$ in topic $t$). 

We thus define a \emph{candidate topical keyphrase} for topic $t$ to be a set of words $p=\{w_1\ldots w_n\}$ which are simultaneously labeled with topic $t$ in at least $\mu$ titles, as discovered by the frequent pattern mining step. We then move on to evaluating the quality of the candidate topical keyphrases in order to rank them within each topic.

\subsection{Ranking Measures and Function}
\label{sec:rank}

As discussed in Section \ref{sec:intro}, a ranking function must compare topical keyphrases with respect to the four criteria of Coverage, Purity, Phraseness, and Completeness. This implies that the function should be able to directly compare keyphrases of mixed lengths, which we refer to as having the \textbf{comparability property}. For example, the keyphrases `classification,' `decision trees,' and `support vector machines' should all be ranked highly in the integrated list of keyphrases for the Machine Learning topic, in spite of having different lengths. 

Traditional probabilistic modeling approaches such as language models or topic models do not have the comparability property. They can model the probability of seeing an n-gram\hide{ for a set $n$} given a topic, but the probabilities of n-grams with different lengths (unigrams, bigrams, etc) are not well comparable. These approaches simply find longer n-grams to have much smaller probability than shorter ones, because the probabilities of seeing every possible unigram sum up to 1, and so do the probabilities of seeing every possible bigram, trigram, etc. But the total number of possible n-grams grows following a power law ($O(v^n)$ where $v$ is the vocabulary size), and therefore ranking functions based on these traditional approaches invariably favor short n-grams. While previous work has used various heuristics to correct this bias during post-processing steps, \cite{Tomokiyo03,Zhao11}, our approach is cleaner and more principled.

We propose a different ranking model which exhibits the comparability property. The key idea is to represent the random event $e_t(p)=$`seeing a phrase $p$ in a random title with topic $t$'. With this definition, the events of seeing n-grams of various lengths in different titles are no longer mutually exclusive, and therefore the probabilities no longer need to sum up to 1.
Formally, the probability of $e_t(p)$ is defined to be $P(e_t(p))=\frac{f_t(p)}{|D_t|}$. In the subsequent sections we define our measures representing the 4 criteria of coverage, purity, phraseness, and completeness using quantities related to this probability. 

\subsubsection{Coverage}
\vspace{-0.1cm}
A representative keyphrase for a topic should cover many documents within that topic. For example, `information retrieval' has better coverage than `cross-language information retrieval' in the topic of Information Retrieval. We directly quantify the coverage measure of a phrase as the probability of seeing a phrase in a random topic-$t$ word set $p^t_d\in D_t$:\\
\vspace{-0.7cm}
\begin{align}
\pi^{cov}_t(p)=P(e_t(p))=\frac{f_t(p)}{|D_t|} 
\end{align}
\vspace{-0.9cm}

\subsubsection{Purity}

A phrase is pure in topic $t$ if it is only frequent in documents about topic $t$ and not frequent in documents about other topics. For example, `query processing' is a more pure keyphrase than `query' in the Databases topic. 

We measure the purity of a keyphrase by comparing the probability of seeing a phrase in the topic-$t$ collection of word sets and the probability of seeing it in any other topic-$t'$ collection ($t'=0,1,\ldots,k$, $t'\ne t$). A reference collection $D_{t,t'}=D_t\cup D_{t'}$ is a mix of of topic-$t$ and topic-$t'$ titles. If there exists a topic $t'$ such that the probability of $e_{t,t'}(p)=$`seeing a phrase $p$ in a reference collection $D_{t,t'}$' is similar or even larger than the probability of seeing $p$ in $D_t$, the phrase $p$ indicates confusion about topic $t$ and $t'$. The purity of a keyphrase compares the probability of seeing it in the topic-$t$ collection and the maximal probability of seeing it in any reference collection:
\begin{align}
\pi^{pur}_t(p)&=\log \frac{P(e_t(p))}{\max_{t'} P(e_{t,t'}(p))} \\
& = \log \frac{f_t(p)}{|D_t|}-\log\max_{t'}\frac{f_t(p)+f_{t'}(p)}{|D_{t,t'}|} \nonumber
\end{align}
\subsubsection{Phraseness}

A group of words should be grouped into a phrase if they co-occur significantly more frequent than the expected co-occurrence frequency given that each word in the phrase occurs independently. \hide{However, high co-occurrence frequency of $n$ words does not necessarily imply the set of $n$ words has good phraseness, since popular words may co-occur often by chance. }For example, while `active learning' is a good keyphrase as in the Machine Learning topic `learning classification' is not, since the latter two words co-occur only because both of them are popular in the topic. 

We therefore compare the probability of seeing a phrase $p=\{w_1\ldots w_n\}$ and seeing the $n$ words $w_1\ldots w_n$ independently in topic-$t$ documents:
\begin{align}
\pi^{phr}_t(p)&=\log \frac{P(e_t(p))}{\prod_{w\in p}P(e_t(w))}\\
&=\log \frac{f_t(p)}{|D_t|}-\sum_{w\in p}\log \frac{f_t(w)}{|D_t|} \nonumber
\end{align}

\subsubsection{Completeness}
A phrase $p$ is not complete if a longer phrase $p'$ which contains $p$ usually co-occurs with $p$. For example, `vector machines' is not a complete phrase but `support vector machines' is, because 'support' almost always accompanies `vector machines'.

We thus measure the completeness of a phrase $p$ by examining the conditional probability of observing $p'$ given $p$ in a topic-$t$ title:
\begin{align}
\pi^{com}_t(p)&=1-\max_{p' \supsetneqq p}P(e_t(p')|e_t(p)) \\
&=1-\max_{w}P(e_t(p\cup \{w\})|e_t(p)) \nonumber \\
&=1-\frac{\max_w f_t(p\cup\{w\})}{f_t(p)} \nonumber
\end{align}

\vspace{-0.2cm}
\subsubsection{Combined Ranking Function} \label{sec:rank_combo}
We combine these 4 measures into a comprehensive function for ranking a topical keyphrase:
\begin{equation}
\label{eq:rank}
r_t(p)=\begin{cases}
0 & \pi^{com}_t\le \gamma \\
\pi^{cov}_t[(1-\omega)\pi^{pur}_t+\omega\pi^{phr}_t](p) & \text{o.w.}
\end{cases}
\end{equation}
where $\gamma,\omega\in [0,1]$ are two parameters.

The completeness criterion is used as a filtering condition to remove incomplete phrases, where $\gamma$ controls how aggressively we prune. $\gamma=0$ corresponds to ignoring the criteria and retaining all \emph{maximal} phrases, where no supersets have the same topical support. As $\gamma$ approaches 1, more phrases will be filtered and only \emph{closed} phrases (where no supersets are sufficiently frequent) will remain. The other three criteria then calculate the ranking score for the keyphrases which pass the completeness filter. 


The coverage criterion is in some sense the most important, since a keyphrase with low coverage will be obviously of low quality. In $r_t(p)$, $\pi_t^{cov}(p)$ is a probability $P(e_t(p)) $ and multiplies both $\pi_t^{pur}$ and $\pi_t^{phr}$. This is desirable because when $P(e_t(p))$ is small, phrase $p$ will have low support, and thus the estimates of purity and phraseness will be unreliable and play a minor role.

The tradeoff between purity and phraseness is controlled by $\omega$. Both measures are log ratios on comparable scales, and can thus be balanced by a weighted summation. As $\omega$ increases we expect more topic-independent but common phrases to be ranked higher. \hide{Though there is technically no positive limit on the value of $\omega$, we restrict $\omega\in [0,1]$ because for our task, we value keyphrases that are pure in their topics.}

The ranking function can also be nicely explained in an information theoretic framework. In fact, the product of coverage and purity,
$\pi^{cov}_t(p)\pi^{pur}_t(p) = P(e_t(p))\log\frac{P(e_t(p))}{P(e_{t,t^*}(p))}$ is equal to the pointwise KL-divergence between the probability of $e_t(p)$ and $e_{t,t^*}(p)$. Pointwise KL-divergence is a distance measure between two probabilities that takes the absolute probability into consideration, and is more robust than pointwise mutual information when the relative difference between probabilities need to be supported by sufficiently high absolute probability. Likewise, the product of coverage and phraseness, $\pi^{cov}_t(p)\pi^{phr}_t(p)$ is equivalent to pointwise KL-divergence between the probability of $e_t(p)$ under different independence assumptions. Therefore, Eq.~(\ref{eq:rank}) can also be interpreted as a weighted summation of two pointwise KL-divergence metrics.


\begin{table*}[t!]
	\renewcommand{\arraystretch}{0.9}
	 \setlength{\tabcolsep}{5pt}
 \centering
  \footnotesize 
    \begin{tabularx}{\linewidth}{ |p{1.6cm}|p{1.6cm}|p{1.6cm}|p{2.9cm}|p{1.7cm}|p{2.1cm}|p{2.5cm}| }

 \toprule 
    
\textbf{kpRelInt\textsuperscript{*}} & \textbf{kpRel} & \textbf{KERT\textsubscript{--cov}} & \textbf{KERT\textsubscript{--pur}} & \textbf{KERT\textsubscript{--phr}} & \textbf{KERT\textsubscript{--com}} & \textbf{KERT}\\
\midrule 
learning & learning & effective & \specialcell{support vector\\machines} & learning & learning & learning \\ 
\arrayrulecolor{Gray}\hline
classification & classification & text & feature selection & classification & \specialcell{support vector\\machines} & \specialcell{support vector\\machines} \\ 
\arrayrulecolor{Gray}\hline
selection & \specialcell{learning\\classification} & probabilistic & \specialcell{reinforcement\\learning} & selection & support vector & \specialcell{reinforcement\\learning} \\ 
\arrayrulecolor{Gray}\hline
models & selection & identification & \specialcell{conditional\\random fields} & feature & \specialcell{reinforcement\\learning} & \specialcell{feature\\selection} \\ 
\arrayrulecolor{Gray}\hline
algorithm & \specialcell{selection\\learning} & mapping & constraint satisfaction & decision & feature selection & \specialcell{conditional\\random fields} \\ 
\arrayrulecolor{Gray}\hline
feature & feature & task & decision trees & bayesian & \specialcell{conditional\\random fields} & classification \\ \arrayrulecolor{Gray}\hline
decision & decision & planning & \specialcell{dimensionality\\reduction} & trees & vector machines & decision trees \\ \arrayrulecolor{Gray}\hline
bayesian & bayesian & set & \specialcell{constraint\\satisfaction problems} & problem & classification & \specialcell{constraint\\satisfaction} \\ 
\arrayrulecolor{Gray}\hline
trees & \specialcell{feature\\learning} & subspace & matrix factorization & \specialcell{reinforcement\\learning} & \specialcell{support\\machines} & \specialcell{dimensionality\\reduction} \\ 
\arrayrulecolor{Gray}\hline
problem & trees & online & hidden markov models & constraint & decision trees & matrix factorization \\ 

\bottomrule 
\end{tabularx}
  \caption{Top 10 ranked keyphrases in the Machine Learning topic by different methods.}
  \label{tbl:compare_methods_top_10}
\vspace{-0.2cm}
\end{table*}

\vspace{-0.1cm}
\section{Experiments} \label{sec:exp}
\vspace{-0.1cm}

We use two collections of content-representative titles in our evaluation. The first - the DBLP dataset - is a set of titles of recently published computer science papers in the areas related to Databases, Data Mining, Information Retrieval, Machine Learning, and Natural Language Processing. These titles come from DBLP\footnote{http://www.dblp.org/}, a bibliography website for computer science publications. 
The second collection - the arXiv dataset - is a sample of titles of physics papers published within the last decade, and labeled by their authors as belonging to the subfields of Optics, Fluid Dynamics, Atomic Physics, Instrumentation and Detectors, or Plasma Physics. This collection of titles comes from arXiv\footnote{http://arxiv.org}, an online archive for electronic preprints of scientific papers.\footnote{Both datasets will be online available} 

Both datasets were minimally pre-processed by removing all stopwords from the titles. After pre-processing, the DBLP dataset contained 33,313 titles consisting of 18,598 unique words, and the arXiv dataset contained 9,722 titles evenly sampled from the specified 5 physics subfields, and consisting of 9,648 unique words.

\subsection{DBLP Dataset Experiments}

We use the DBLP dataset to evaluate the ability of our method to construct topical keyphrases that appear to be high quality to human judges, via a user study. We will first describe the methods we used for comparison, and then present a sample of the keyphrases actually generated by these methods and encountered by participants in the user study. We then explain the details of our user study, and present quantitative results. 

\subsubsection{Methods for Comparison}\label{subsec:methods_compare}

We use background LDA introduced in \secref{sec:clustering} for the word clustering step in order to create input for all the methods that we compare. We resort to a Newton-Raphson iteration method~\cite{Minka00} to learn the hyperparameters, and empirically set $\lambda = 0.1$, which leads to generally coherent results for our dataset.\footnote{The learned $\alpha=1.0$ is smaller than the typical setting due to the nature of our short text, and $\beta=0.07$ is larger because in our dataset, different topics often share the same words.} 

To evaluate the performance of KERT, we implemented several variations of the function, as well as two baseline functions. The baselines come from Zhao et al ~\shortcite{Zhao11}, who focus on topical keyphrase extraction in microblogs, but claim that their method can be used for other text collections. We implement their two best performing methods: kpRelInt* and kpRel.\footnote{Their main ranking function kpRelInt considers the heuristics of phrase interestingness and relevance. As their interestingness measure is represented by re-Tweets, a concept that is not appropriate to our dataset, we reimplement the interestingness measure to be the relative frequency of the phrase in the dataset instead, and we therefore refer to our reimplementation as kpRelInt\textsuperscript{*}. kpRel considers only the relevance heuristic.} We also construct variations of KERT where the keyphrase extraction steps are the same, but each of the four ranking criteria is ignored in turn. We refer to these versions as KERT\textsubscript{--cov}, KERT\textsubscript{--pur}, KERT\textsubscript{--phr}, and KERT\textsubscript{--com}, respectively.



These variations nicely represent the possible settings for the parameters $\gamma \in [0,1]$ and $\omega  \in [0,1]$, which are described in Section~\ref{sec:rank_combo}. In KERT we set $\gamma=\omega=0.5$. KERT\textsubscript{--com} sets $\gamma=0$ to demonstrate what happens when we retain all \emph{maximal} phrases. As $\gamma$ approaches 1, more phrases will be filtered but a very small number of \emph{closed} phrases (no supersets are frequent) will not be. KERT\textsubscript{--phr} sets $\omega=0$ and KERT\textsubscript{--pur} sets $\omega=1$, which demonstrates the tradeoff between ignoring phraseness for the sake of maximizing purity, and ignoring purity to optimize for phraseness, respectively.

\subsubsection{Qualitative Results}

Table~\ref{tbl:compare_methods_top_10} shows the top 10 ranked topical keyphrases generated by each method for the topic of Machine Learning. kpRel and kpRelInt\textsuperscript{*} yield very similar results, both clearly favoring unigrams. However, kpRel also ranks several keyphrases highly which are not very meaningful, such as `learning classification' and `selection learning.' Removing coverage from our ranking function yields the worst results, confirming the intuition that a high quality keyphrase must at minimum have good coverage. Without purity, the function favors bigrams and trigrams that all appear to be very meaningful, although several high quality unigrams such as `learning' and `classification' no longer appear. Removing phraseness, in contrast, yields meaningful unigrams but very few bigrams, and looks quite similar to the kpRelInt\textsuperscript{*} baseline. Finally, without completeness, phrases such as `support vector' and `vector machines' are ranked highly, although they should not be, as both are sub-phrases of the high quality trigram `support vector machines.'

\subsubsection{User Study and Quantitative Results}

To quantitatively measure keyphrase quality, we invited people to judge the generated topical keyphrases generated by the different methods. Since the DBLP dataset generates topics in computer science, we recruited 10 computer science graduate students - who could thus be considered to be very knowledgeable judges - for a user study. 
\hide{We first introduced the participants to the concept of topical keyphrases, and suggested that the criteria of coverage, purity, phraseness, and completeness should guide their evaluation. }{We generated 5 topics from the DBLP dataset and found 4 of them were clearly interpretable as Machine Learning, Databases, Data Mining, and Information Retrieval.}For each of the four topics, we retrieved the top 20 ranked keyphrases by each method. These keyphrases were gathered together per topic and presented in random order, and users were asked to evaluate the quality of each keyphrase on a 5 point Likert scale.

To measure the performance of each method given the user study results, we adapt the \textbf{nKQM@K measure} (normalized keyphrase quality measure for top-K phrases) from \cite{Zhao11}, which is itself a version of the \emph{nDCG} metric from information retrieval \cite{Jarvelin02}. We define nKQM@K for a method \emph{M} using the top-K generated keyphrases as:

\vspace{-0.6cm}
\begin{equation*} \label{eq:nKQM}
nKQM@K = \frac{1}{T}\sum_{t=1}^{T} \frac{\sum_{j=1}^{K} \frac{score_{aw}(M_{t,j})}{log_2(j+1)}} {IdealScore_K}
\end{equation*}
\vspace{-0.2cm}

Here $T$ is the number of topics, and $M_{t,j}$ refers to the $j^{th}$ keyphrase generated by method $M$ for topic $t$. Unlike in \cite{Zhao11}, we have more than 2 judges, so we define $score_{aw}$ as the \emph{agreement-weighted} average score for the $M_{t,j}$ keyphrase, which is a weighted mean of the judges' score by the weighted Cohen's $\kappa$. This gives a higher value to a keyphrase with scores of (3,3,3) than to one with scores of (1,3,5), though the average score is identical.  Finally, $IdealScore_K$ is calculated using the scores of the top K keyphrases of all judged ones\hide{, and is therefore the same value for every method. We thus consider the combined set of the top K keyphrases generated by all the methods to be the complete list of ranked keyphrases in that topic, and treat the agreement-weighted average scores as rankings for all the keyphrases}.


\renewcommand{\arraystretch}{1.0}
 \setlength{\tabcolsep}{6pt}
\begin{table}[h!]
\centering
  \small
    \begin{tabular}{|l||c|c|c|c|}
    \toprule
\textbf{Method} & \textbf{nKQM\textsubscript{@5}} & \textbf{nKQM\textsubscript{@10}} & \textbf{nKQM\textsubscript{@20}}   \\ 
\midrule

\textbf{KERT\textsubscript{--cov}} 	& 0.2605 & 0.2701 & 0.2448 \\
\textbf{kpRelInt\textsuperscript{*}}	& 0.3521 & 0.3730 & 0.3288 \\
\textbf{KERT\textsubscript{--phr}} 	& 0.3632 & 0.3616 & 0.3278 \\
\textbf{kpRel}				& 0.3892 & 0.4030 & 0.3767 \\
\textbf{KERT\textsubscript{--com}} & \textbf{0.5124} & \textbf{0.4932} & \textbf{0.4338} \\
\textbf{KERT} 				& \textbf{0.5198} & \textbf{0.4962} & \textbf{0.4393} \\
\textbf{KERT\textsubscript{--pur}} 	& \textbf{0.5832} & \textbf{0.5642} & \textbf{0.5144} \\

    \bottomrule
    \end{tabular}
  \caption{nKQM@K values for different methods (ordered by their performance from low to high)}
  \label{tbl:compare_methods_nKQM}
\end{table}

Table \ref{tbl:compare_methods_nKQM} compares the performance across different methods. The top performances are clearly variations of KERT with different parameter settings. As expected KERT\textsubscript{--cov} exhibits the worst performance. The baselines perform slightly better, and it is interesting to note that kpRel, which is smoothed purity, performs better than kpRelInt\textsuperscript{*}, and even slightly better than KERT\textsubscript{--phr}. This is because kpRelInt\textsuperscript{*} adds in a measure of \emph{overall} keyphrase coverage in the entire collection, which hurts rather than helps for this task. Removing completeness appears to have a very small negative effect, and we hypothesize this is because high-ranked incomplete keyphrases are relatively rare, though very obvious when they do occur (e.g. `vector machines'). KERT\textsubscript{--pur} performs the best - which may reflect human bias towards longer phrases - with an improvement of at least 50\% over the kpRelInt\textsuperscript{*} baseline for all reported values of K.

\vspace{-0.2cm}
\subsection{arXiv Dataset Experiments}

We use the arXiv dataset, which contains labeled titles, to explore which method maximizes the mutual information between phrase-represented topics and titles. As the collection has 5 categories, we set T=5.

For each method, we do multiple runs for various values of K (the number of top-ranked phrases per topic considered), and calculate the mutual information $MI_{K}$ for that method as a function of K. To calculate $MI_{K}$, we label each of the top $K$ phrases in every topic with the topic in which it is ranked highest. We then check each paper title to see if it contains any of these top phrases. If so, we update the number of events ``seeing a topic t and category c'' for $t=1\ldots T$, with the averaged count for all those labeled phrases contained in the title; otherwise we update the number of events ``seeing a topic t and category c'' for $t=1\dots T$ uniformly, where c is the Primary Category label for the paper title in consideration. Finally, we compute the mutual information at K:

\vspace{-0.2cm}
\begin{equation}
	MI_K = \sum_{t,c} p(t,c)\ log_2\frac{p(t,c)}{p(t)p(c)} \nonumber
\end{equation}

\begin{figure}[h!]
 \centering
\includegraphics[width=0.52\textwidth]{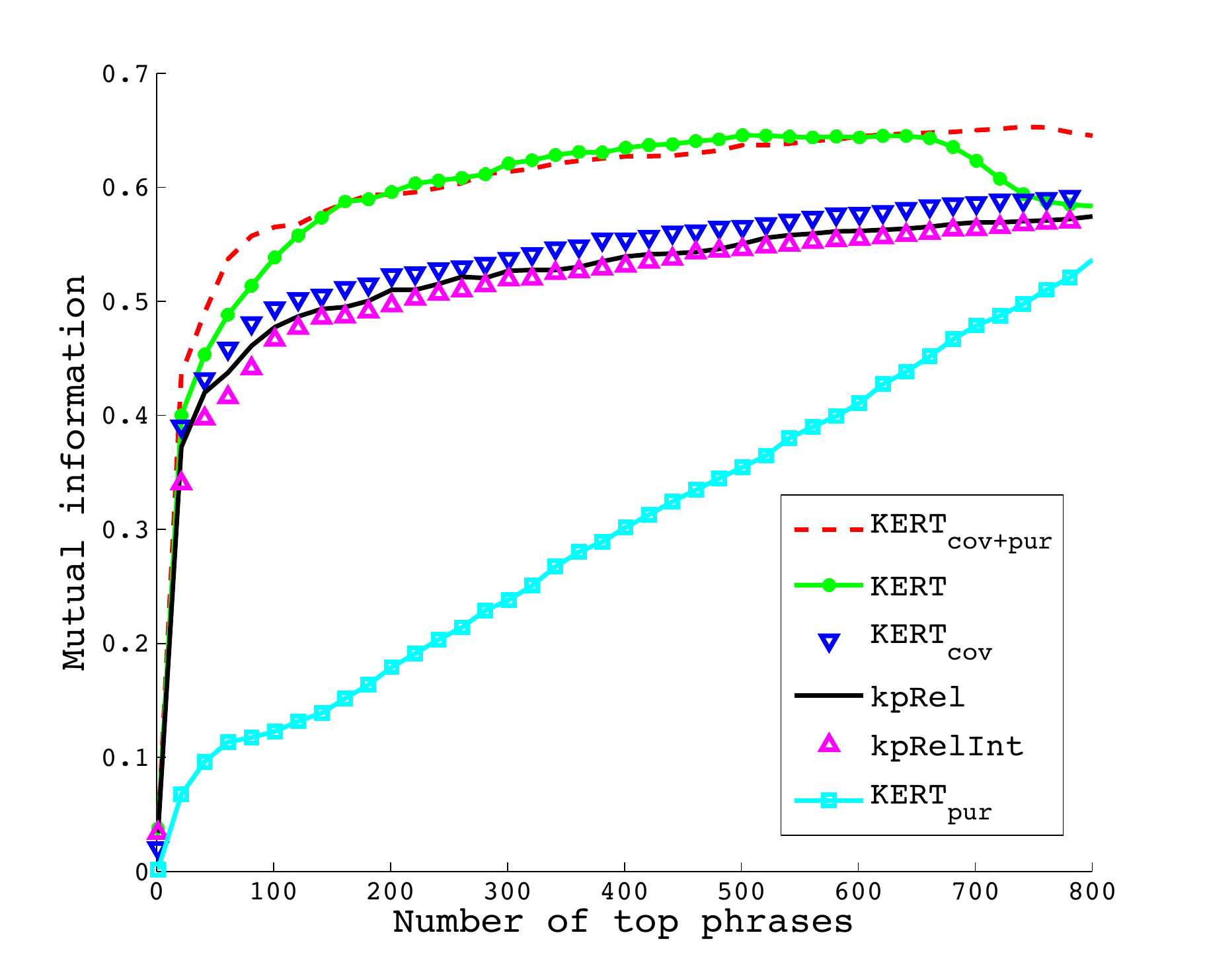}
\vspace{-0.8cm}
\caption{MI\textsubscript{K} values for various methods (methods in legend are ordered by performance, high to low)}
 \label{fig:mi_graph}
 \vspace{-0.5cm}
\end{figure}

We compare the baselines (kpRelInt\textsuperscript{*} and kpRel), $KERT$, and variations of $KERT$ where only coverage ($KERT_{cov}$), only purity ($KERT_{pur}$), and only coverage and purity ($KERT_{cov+pur}$) are used in the ranking function. We feed them the same input by background LDA with the same parameter settings as discussed above. Figure \ref{fig:mi_graph} shows $MI_K$ for each method for a range of K.

It is clear that for MI\textsubscript{K}, coverage is more important than purity, since $KERT_{pur}$ is by far the worst performer. Both baselines perform nearly as well as KERT\textsubscript{cov}, and all are comfortably beaten by KERT\textsubscript{cov+pur} ($>20\%$ improvement for $K$ between 100 and 600), which uses our coverage and purity measure. It is interesting to note that adding in the phraseness and completeness measures yields no improvement in MI\textsubscript{K}. However, the experiments with the DBLP dataset demonstrate that these measures are very helpful in the eyes of expert judges. In contrast, while MI\textsubscript{K} is definitely improved with the addition of the purity measure, people prefer for it to be removed. Although we outperform other approaches in both evaluations, these observations show interesting differences between theory-based and human-based evaluation metrics.


\vspace{-0.1cm}
\section{Conclusion}\label{sec:conc}

In this work we introduce KERT (Keyphrase Extraction and Ranking by Topic), a framework for the automatic extraction and ranking of topical keyphrases using collections of content-representative document titles. Unlike existing techniques, our phrase-centric approach is able to construct candidate topical keyphrases in such a way as to allow our ranking function to directly compare the quality of keyphrases of different lengths. Our method yields high quality topical keyphrases, with over 50\% improvement over a baseline method according to human judgement and over 20\% improvement according to mutual information. 
In the future we would like to further explore why human judgement appears to be consistently biased against the purity criteria, in contrast to quantitative metrics such as mutual information. We are also interested in extending our approach to working with longer texts.



\bibliographystyle{acl2012}
\bibliography{KERT}  

\end{document}